\documentclass{article}
\usepackage[utf8]{inputenc}

\title{Combined Learning of Neural Network Weights for Privacy in Collaborative Tasks}
\author{Aline R. Ioste \and Alan M. Durham \and Marcelo Finger}

\date{
	Universidade de  São Paulo - Brazil
}
\usepackage{url}
\usepackage{a4wide}
\usepackage{graphicx}
\usepackage{algorithm}
\usepackage{amsmath,amsfonts}
\usepackage[noend]{algpseudocode}
\usepackage{caption}
\usepackage[shortlabels]{enumitem}
\usepackage[colorlinks=true,citecolor=blue]{hyperref}
\usepackage[capitalize]{cleveref}
\DeclareUnicodeCharacter{2212}{-}

\usepackage{todonotes}

\begin{document}

\maketitle
\begin{abstract}
We introduce CoLN, Combined
Learning of Neural network weights, a novel method to securely combine Machine Learning models over sensitive data with no sharing of data. With CoLN, local hosts use the same Neural Network architecture and base parameters to train a model using only locally available data. Locally trained models are then submitted to a combining agent, which produces a combined model. The new model's parameters can be send back to hosts, and can then be used as initial parameters for a new training iteration. CoLN is capable of combining several distributed neural networks of the same kind, but is not restricted to any single neural architecture. In this paper we detail  the combination algorithm and present experiments with feed-forward, convolutional and recurrent Neural Network architectures, showing that the CoLN combined model approximates the performance of a hypothetical ideal centralized model, trained using the combination of the local datasets.
CoLN can contribute for secure collaborative research, as required in the medical area, where privacy issues preclude data sharing, but where the limitations of local data demand information derived from larger datasets. 
\end{abstract}

\section{Introduction}

During the standard training of neural network models, it is usually assumed that all data is readily available to be used by the  application.
However, this assumption is not valid when handling sensitive data that is distributed over several hosts with  access restrictions~\cite{de2012data}. Existing data protection laws, in particular the  General Data Protection Regulation (GDPR)~\cite{10.5555/3002861}, limit data sharing and, as a consequence, the potential size of the training sets, which in turn can reduce the accuracy of the models.  In fact,  several  applications of  Deep Learning in Health have already pointed to the conflict between the need to collaborate and the need to respect privacy constraints~\cite{ravi2016deep, ning2020open, dias2020deeplms}. The consequence is a restriction on  collaborative research across international borders,  when  private patient data is involved~\cite{bennett1992regulating,patil2014big,price2019privacy}.  

The machine learning community has proposed several approaches to provide data protection while still using centralized training, including: data anonymization,  differential privacy \cite{abadi2016deep, dwork2008differential}, homomorphic encryption to data cryptography  \cite{graepel2012ml,wood2019homomorphic}, and combined data anonymization with random noise generation, enhanced with homomorphic encryption \cite{sharma2020use}. However, these approaches are not always feasible. First, anonymization may still expose personal information \cite{rocher2019estimating}; second, some approaches may reduce the final performance of the machine learning method \cite{graepel2012ml, abadi2016deep}; third, different privacy and security restrictions in the data may still  preclude exporting training data to centralized locations, specially when this involves data crossing international borders \cite{russom2011big,de2012data}.

Decentralized learning addresses these challenges \cite{williams2020defining,montes2019distributed,helbing2019will, ahlbrandt2014balancing,ras2018explanation}. In a decentralized learning approach, separate models are trained locally by separate hosts, and a combined model is produced by using the parameters and metadata of the local models. Such an approach is  more privacy compliant,  as no data is shared and it is not possible to reconstruct input data from the parameters of locally trained models~\cite{angles2019privacy,soliman2020decentralized}. 

In the context of decentralized learning, one of the most popular approached is \textit{Federated Learning}, initially proposed in the Federated Average (FedAvg) algorithm, where local parameters are averaged to obtain the centralized values~\cite{mcmahan2017communication,konevcny2016federated}.

Several variations of this technique were proposed:
\begin{itemize}
    \item \textit{Federated Proximal} (FedProx)~\cite{li2018federated},  an approach designed to optimize network traffic. This approach was proposed as a federated optimization structure for heterogeneous networks. This techique limits the impact of local updates, restricting them to be close to the current model.

    \item \textit{Agnostic Federated Learning (AFL)}~\cite{mohri2019agnostic}, an approach designed to cope with with non-uniform distribution of data points which is  agnostic with respect to data distribution, 
    assuming that global loss is an unknown \textit{convex combination} of local data. 
    AFL uses a stochastic optimization process which applies a  regularization that provides learning guarantees.  
    
    \item \textit{Probabilistic Federated Neural Matching (PFNM)}~\cite{yurochkin2019bayesian} proposes another form of parameter combination that identifies subsets of neurons in each of the local models that match neurons in other local models, and then  combine the matched neurons to form a global model. However, the method only works with simple network architectures, without applicability to to Convolutional Neural Networks (CNNs) and Recurrent Neural Networks (RNNs).

    \item \textit{Federated Matched Averaging (FedMA)}~\cite{wang2020federated} extends PFNM ideas to CNNs and LSTMs. FedMA is a layers-wide federated learning algorithm that employs Bayesian non-parametric methods to adapt to heterogeneity in the data. 
\end{itemize}

We present  Combined Learning of Neural network weights (CoLN), a new Distributed Learning approach to combine parameter information inferred autonomously by several distributed hosts engaged in collaborative work.  Only local model parameters and information on local dataset sizes are transmitted over the network by the collaborating hosts. The method  avoids traffic issues present in some distributed learning algorithms, with very little information being exchanged in the network and a reduced number of steps required in the process. More important, the method is very resilient to variations in the local data distributions and dataset sizes. 

CoLN achieves the goal of obtaining, from a set of models locally trained on  distinct datasets, a combined model that approximates the performance of a hypothetical ideal centralized model, which  would be the result of using the combined local datasets for training a single model. 

\newpage
\section{Methods}\label{sec:methodology}
 CoLN is an interactive approach that, at each round, first performs training with the models initialized by the same set of parameter values, and then compute a new set of parameters combining the parameters resulting from each local training. 
For the combination phase we use a linear combination of the local values that avoids catastrophic forgetting and, at the same time, ensures generalization.

\subsection{\textbf{Basic architecture}}

We assume there are $H$ \textit{host sites}, each with a distinct  data set and with a  \textit{host model} trained using the local data.  All sites should use the same Neural Network architecture as the basic model and, before training, are initialize it with the same parameters.

Model parameters consist of weights, biases and hyper parameters. In this view a layer is viewed as an individual matrix or tensor, whose parameters are called simply \emph{weights}; biases are assumed to be elements of layers like any other weights in the network.  

For the method's presentation, we refer to a \textit{synchronizing site} in charge of combining the locally trained host models. However, the parameter combination process to obtain the combined model can be performed in one of the hosts, or in parallel by all local hosts. 

The CoLN central hypothesis is that, when we have H local hosts each with W weights, we can combined the learning of all hosts usint  a linear transformation on local host weights ($w^h_i$): 
\begin{align}\label{eq:CoLN}
    \overline{w}_i &= \sum_{h=1}^H \alpha^h w^h_i + \beta_i , \qquad\qquad 1 \leq i \leq W
\end{align}

A central hypothesis of this work, which we tried to empirically validate,  is that a linear combination is a good way to combine network weights.  Intuitively, linear combinations are a well-founded and simple  method, which  allows for quite flexible and expressive combinations based on the coefficients $\alpha$ and $\beta$. The coefficients $\alpha$ and $\beta$ are refined below with reference to \emph{network layers} in~\cref{sec:coeff}.  The value of $\alpha^h$ reflects the relevance of host $h$ relative to other  hosts; the value of $\beta_i$ is empirically determined based on a notion of distance between the weights of host $h$ and a hypothetical centralized model. Empirically, the linear combination detailed in~\cref{sec:coeff} satisfies the following characteristics: 

\begin{enumerate}[($i$)]
    \item The composed model is  closer to the centrally trained model $w^*$, that is, $\textit{dist}(\overline{w},w^*) < \textit{dist}({w}^h,w^*)$ for every $h \in [1,H]$.  The distance is measured by~\cref{eq:dpmpl0}, but it could be understood in terms of improved accuracy, namely that the combined model is closer to the accuracy of a hypothetical centralized model than any of the local hosts.
    \item We can re-train the local parameters using initial weights $\overline{w}$ 
    and reiterate the process, obtaining  obtaining $\overline{\overline{w}}$, a better approximation of $w^*$-performance, that is,  $\textit{dist}(\overline{\overline{w}},w^*) < \textit{dist}(\overline{w},w^*)$.
    
\end{enumerate}

The first characteristic would give us the possibility of producing a strongly-decoupled distributed learning method with low communication overhead, in which each host starts with the same initial model $w^0$  and, with its own dataset, autonomously produces a model $w^h$ which is sent only  to a synchronizer, which computes $\overline{w}$. The second characteristic would ensure that  this process could be iterated to obtain an even better model.

It is important to note that, different than that other approaches,  we do not require the linear coefficients to be convex, that is, for a fixed weight $i$, $\alpha^1, \ldots, \alpha^H , \beta_i$ are not required to add up to 1. In fact,  previous experiments indicated that   convexity does not guarantee characteristics ($i$) and ($ii$), suggesting that the centralized optimal model lies outside of the convex hull of the local optimum models.  This fact should not be a surprise. If the centralized model resulting from an optimization using all the data were obtainable as a convex combination of host models, this would indicate a linear nature of the neural models, contrary to the existence of non-linear activation functions in almost all non-trivial neural models.

  In CoLN, model combination proceeds in a sequence of rounds, whose main characteristics are as follows.

\begin{enumerate}
    \item \textbf{Initialization:} At the beginning of each round all hosts use \emph{the same} set of parameters. This is a synchronization point.  In the first round, the initial weights could be generated randomly and then distributed, or obtained  from a previously trained model.  In subsequent rounds, the initial weights of the models are the result of the CoLN combination of previous round.

    \item \textbf{Training:} Each host trains its model based only on locally available data, independently from all other hosts. The number of epochs used by each host site may vary; locally available datasets can be of different sizes and present different distributions.
    
    \item \textbf{Synchronizing:} After local training, local model weights are exported to a synchronizing site, where a synchronized model is computed using the linear transformation. The round finishes with the redistribution of the synchronized model weights back to the hosts.

\end{enumerate}

\subsection{Combination Coefficients}
\label{sec:coeff}

We now will refine and detail the combination method described in equation ~\eqref{eq:CoLN}. For this we need to associate individual weights with a   \emph{layer} in the model. 

The linear combination have the following formal parameters:

\begin{itemize}
    \item $L$, the number of layers of the common architecture;
    \item $M_\ell$, the number of weights and biases in layer $\ell$, $\ell \in [1, L]$ of the common architecture;
    \item $H$, the amount of host models;
    \item $T_h$, the size of the training corpus at host site $h$; the total size of training data is $T= \sum_{h=1}^H T_h$; 
    \item $r_h$, the relative size of the training corpus at $d$ with respect to the total size, $r_h = \frac{T_h}{T}$.
    \item $w_{i,\ell}^h$, the value of $i$th weight in  layer $\ell$ for host model $h$, $1 \leq i \leq m_\ell$, $1 \leq h \leq H$.
    \item $\overline{w}_{i,\ell}$, the value of $i$th weight in  layer $\ell$ in the synchronized model.
\end{itemize} 

Now the can refine the combination method~\eqref{eq:CoLN} to:

\begin{align}\label{eq:CoLNbyLayer}
    \overline{w}_{i,\ell} &= \sum_{h=1}^H \alpha^h w^h_{i,\ell} + \beta_{i,\ell} , \qquad\qquad 1 \leq i \leq M_\ell, 1 \leq \ell \leq L
\end{align}

Equation~\eqref{eq:CoLNbyLayer} involves both layer independent \emph{linear combiners} $\alpha^h$ and layer dependent \emph{shift values} $\beta_{i,\ell}$; described below.

\textbf{Determining Combination Rate $\alpha^h$}

\begin{figure}[hbt!]
  \centering
  \includegraphics[width=3.0in]{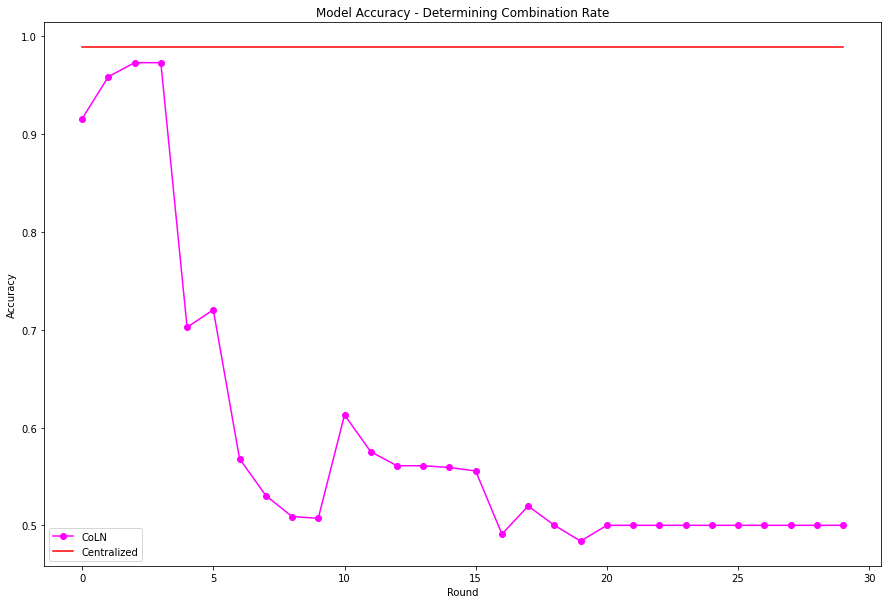}
  \caption{Combinated accuracy in COVID-19 data with  $\mathrm{\alpha}^{h} = 1 + rh, h = 1,2$.}
  \label{fig:cm-hypothesis1}
\end{figure}

The first experimental study aimed at evaluating the format of the linear combination and its
hyper-parameters. For that, we used the COVID-19 dataset with the aim of predicting the
presence of COVID-19 in patients based on a series of other laboratory exams. From the original
dataset, two hosts were artificially built, one with 100\% positive COVID-19 cases and one with 100\% negative cases, and a balanced global test with 50\% positive and 50\% negative cases. The
goal of using models with high discrepancy is to have a clear experimental view of accuracy gain
by combining the 50\% accuracy local models and determining the best linear combiners to achieve a combined model that can achieve a high rating. 
Note that at this point, we did not compare with a centralized model trained on all data which, for the record, reaches 99.64\%.
The initial idea was to study the feasibility of initializing identically two host models locally
training them and then to combine them using linear combiner value of $\alpha^h$ = 1 + $r_h$. In the model exposed in Equation~\eqref{eq:CoLNbyLayer} to \eqref{eq:betacoef}, this corresponds to c = 1 expanded only for the first two terms of its Taylor-series format. Both models were trained with the number of epochs necessary to achieve the best local accuracy. The models are submitted to several rounds of CoLN-combination
and the results are shown in figure \ref{fig:cm-hypothesis1}.

\begin{figure}[hbt!]
  \centering
  \includegraphics[width=3.0in]{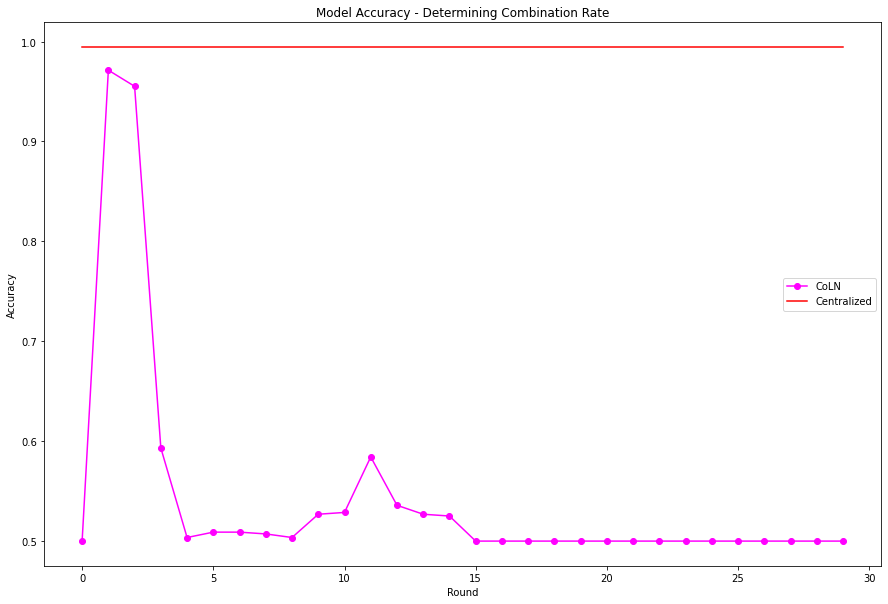}
  \caption{Combinated accuracy in COVID-19 data with  $\mathrm{\alpha}^{h} = \mathrm{e}^{c.r_h}$, $c =+1, h = 1,2$.}
  \label{fig:cm-hypothesis2}
\end{figure}

These initial combined accuracy obtained in the first round is 94,803\%, and it keeps improving
for the next three rounds to 97.31\%, but then it starts decreasing and never recovers until, at
round 24, it reaches about 50\%, which was the accuracy of each model trained separately.
At this point, we had two alternatives, either establishing a method to stop at the appropriate
combination round, which in this case was round 4, or to devise a method which would not suffer from this form of decay. We decided for the second alternative, exploring linear combiners of the
form $\alpha^h= \mathrm{e}^{c \cdot r_h}$. For the case c = 1, the results obtained are shown in figure \ref{fig:cm-hypothesis2}.

\begin{figure}[hbt!]
  \centering
  \includegraphics[width=3.0in]{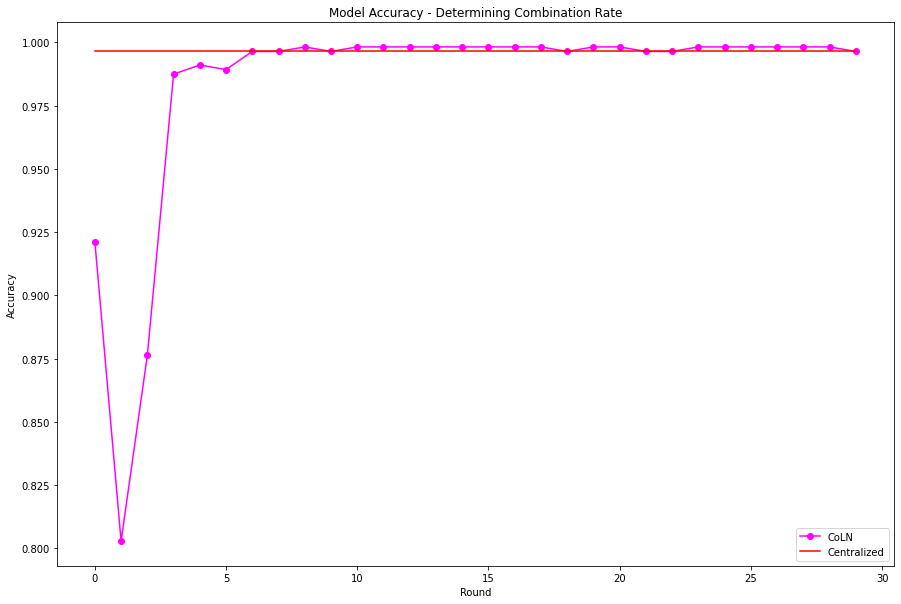}
  \caption{Combinated accuracy in COVID-19 data with $\alpha^h= \mathrm{e}^{c \cdot r_h}$, $c =+1/c = {(−1)},h = 1,2$.}
  \label{fig:cm-hypothesis3}
\end{figure}

In this case, in only three rounds we reached maximum combined accuracy of 97.32\%, just like
before, but the decay in figure \ref{fig:cm-hypothesis2} was very much like that in figure \ref{fig:cm-hypothesis1}, and we noted that both the
average size of the weights as well as their standard deviation were increasing at each round, much
like the phenomenon of exploding gradients. So, we searched for methods to avoid this explosion, or somehow revert it. Employing a value of c = −1 in $\alpha^h= {e}^{c \cdot r_h}$ led to a degradation of performance, that is, a total failure of combination, but then we had the idea of alternating
c = +1 with c = -1 in consecutive rounds, and the results of this alternation are shown in figure \ref{fig:cm-hypothesis3}.

\begin{figure}[hbt!]
  \centering
  \includegraphics[width=3.0in]{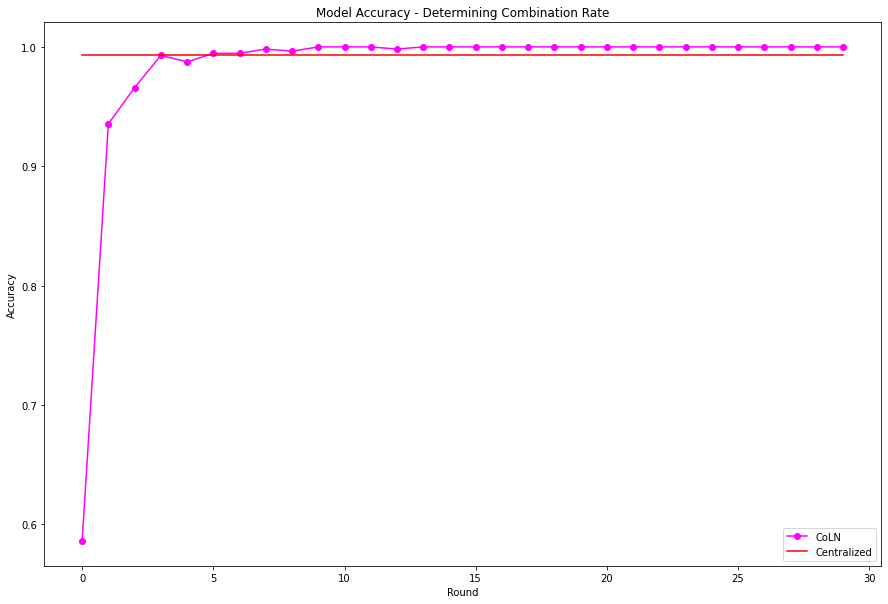}
  \caption{Combinated accuracy in COVID-19 data with $\mathrm{\alpha}^{h} = \mathrm{e}^{c.r_h}$, $c =10^{(-3)}$, h = 1,2}
  \label{fig:cm-hypothesis4}
\end{figure}

Starting from a random model, we accuracy of 66.129\% in the first iteration, 64.51\% in the second,
98.92\% in the third, 98.74\% in the fourth and 99.10\% in the fifth. We can see that, in the alternative case, the phenomenon of sharp accuracy decay in precision is not present. In this setting, we can decide to halt the combination rounds once the accuracy increase between rounds stays within some interval (another hyper-parameter).  
However, the alternation process is inefficient and computationally expensive. In fact, as seen in figure \ref{fig:cm-hypothesis3}, when we alternate weight expansion rounds $c ={1}$ with weight contracting rounds $c ={-1}$, the latter combination never increases accuracy and is employed only to avoid weight degradation, so that half rounds are not moving the combination toward maximum accuracy; but once this point is reached, it stays there. We still do not consider it as the best solution, and investigate other combination rate values.
In traditional gradient descent backpropagation method, the learning rate allows one to in-
corporate a fraction of the gradient at every training epoch. When the learning rate is large, the
learning process may diverge, so usually small values of learning rate are used for better results.
In analogy to this case, we experimented with small combination rates, without alternation, to
avoid weight decay. In that spirit, we explored linear combiners of the form $\alpha^h= \mathrm{e}^{c \cdot r_h}$,
with $c =10^{(-3)}$; the results obtained are shown in figure \ref{fig:cm-hypothesis4}

Combining the network weights in small hops avoids the degradation of the central model without requiring expensive alternations, as seen in figure \ref{fig:cm-hypothesis4}. Starting from a random model, we accuracy of 49,64\% in the first iteration, 78,32\% in the second, 97,85\% in the third, 99,28\% in the
fourth and 100,000\% in the fifth.
In the following, we fix CoLN combination rate hyperparameter at $c=10^{(−3)}$ employing linear combiners of the form equation \ref{eq:alfacoef}. Experiments showed that it is possible to start the combination process with an already trained model by some host, with faster convergence. However, to avoid searching for a volunteer to start the process, we always start with a random model.

Linear CoLNbyLayerombiners are host-dependent exponential factors based on the relative size of the host's dataset:
\begin{align}\label{eq:alfacoef}
	\alpha^h= \mathrm{e}^{c \cdot r_h}
 \end{align} 
where $r_h$ is the host's relative dataset size; and $c$ is a hyper-parameter, the \emph{combination rate}. 

In the empirical validation of the linear combination of layer weights, several alternatives for the value of $\alpha^h$ were considered.  Due to the non-linear nature of neural networks, a convex combination of weights was ruled out, and a  we initially considered using the value  $\alpha^h_1= 1+r_h$. Then we realized that $1+r_d$ are just the first two terms of the Taylor-series expansion of the more general formula {\color{blue}$\alpha^h_2= \mathrm{e}^{c.r_h}$,} and noted that this form for the  linear combiner leads to a faster convergence of the CoLN algorithm; note that $\alpha^h_2$ corresponds to \eqref{eq:alfacoef} with combination rate $c=+1$.

In the supplementary material we describe  the series of  experiments on different families of coefficients $\alpha^1, \ldots, \alpha^H, \beta$, which led to CoLN's final linear combination coefficients. In fact, these experiments 
showed that both $\alpha^h_1$ and $\alpha^h_2$ led to an explosive growth of weight values  after a few combination rounds, which in turn leads to a sharp drop in the combined accuracy, a phenomenon akin to the exploding gradients in Recurrent Nerual Networks~\cite{PMB2013}.  

After experimenting with alternating values of $c=+1$ and $c=-1$ in successive combination rounds, we settled for values of $c \in [10^{-6}$,$10^{-3}]$, which, in our experiments, avoided the weight size explosion phenomenon (see supplementary material).

The computation of shift values employs two auxiliary functions.  The first is a measure of a distance between the specific weight and its counterparts in the other host models, as given by:

\begin{align}\label{eq:dpmpl0}
    \textit{WeightDistance}(i,\ell) &= \sqrt{\sum_{j=1}^{H-1} \sum_{k=j+1}^{H} \left(\strut
       w_{i,\ell}^j \cdot r_j  - w_{i,\ell}^k \cdot r_k \right)^2}
\end{align}

\noindent According to~\cref{eq:dpmpl0}, all weights at host $h$ are scaled by the data size rate $r_h$ prior to computing a summation of all squared euclidian distances between all scaled layer weights, thus producing $\textit{WeightDistance}(i,\ell)$.

The second is a layer distance:

\begin{align}\label{eq:dmpl1}
	\textit{LayerDistance}(\ell) &=\frac{ \sqrt{ \sum_{i=1}^{M_\ell} \sum_{j=1}^{H-1} \sum_{k=j+1}^{H} (w_{i,\ell}^j  - w_{i,\ell}^k)^2}}{M_\ell}
\end{align}
\noindent

The $\textit{LayerDistance}(\ell)$ computes a ponderation over all weights of the sum of the  squared distances between all layer weights.  The idea is to produce a shift for all weights in a layer whose distance to the layer is below average.

We can now define $\beta_{i,\ell}$, the shift value for weight $i$ in layer $l$:

 \begin{align}\label{eq:betacoef}
     \beta_{i,\ell}= \left\{
      \begin{array}{ll}
           \textit{WeightDistance}(i,\ell) , 
           &\textrm{if }  WeightDistance(i,\ell) < LayerDistance(\ell)\\
           0 ,
           &\textrm{otherwise}
      \end{array}
      \right.
 \end{align}

Note that there is a shift only in the cases where the weight distance is smaller than the layer distance; the shift is used to compensate cases in which weights are negligible with respect to other weights in the same layer.

CoLN is summarized by algorithm \cref{alg:euclid}.

\begin{algorithm}[H]
\caption{CoLN}\label{alg:euclid}
\textbf{Input} A. A description of the common model consisting of:  (i)$L$ layers;(ii)$m$, a vector containing  the number of weights per layer. B. The weights of all host models:  $w_{i,\ell,d}, i \in [1,m[\ell]], d \in [1,D], \ell \in [1,L]$; C.  $r$: vector of training data size rates of  per host site.
\\
\textbf{Output} Weights of the combined model \textit{$\overline{W}$}

\begin{algorithmic}[1]
\Procedure{CoLN}{$D,L,m,w,r$}
    \For{$\ell = 1$ to $L$}
        \State Compute $\textit{LayerDistance}(\ell)$
        \For{$i = 1$ to $m[\ell]$}
            \State Compute $\textit{WeightDistance}(i,\ell)$
            \For{$d = 1$ to $D$}
                \State $\overline{w}[i,\ell] \!\!= {\displaystyle \sum_{d=1}^D w[i,\ell,d]* \mathrm{e}^{c \cdot r_h}}$
            \EndFor
            \If {($\textit{WeightDistance}(i,\ell) \leq \textit{LayerDistance}(\ell)$)}
                \State $\overline{w}[i,\ell]\ +\!\!= \textit{WeightDistance}(i,\ell)$
            \EndIf        
        \EndFor
    \EndFor
    \State \Return $\overline{W}$
\EndProcedure
\end{algorithmic}
\end{algorithm}

\section{Results}
\label{sec:exp}

The methods and parameters of CoLN are detailed in~\cref{sec:methodology}. All experiments were performed with the method's hyper-parameters determined in a series of calibration experiments described in the supplementary material.

We have applied CoLN to two sets of  experiments. The aim of the first set was to estimate the accuracy and robustness of CoLN for different types of neural network architectures (Multi-layer feed forward, Convolutional, Recurrent LSTM), different number of local hosts and distribution biases.
The aim of the second set of experiments was to compare CoLN's performance with Federated learning approaches. 

In all experiments,   we also trained a ``centralized model'', using the combination of all local datasets. The performance of this model was used as the ``golden standard'' for comparison. In all cases models were initialized with the same random values before the first training.

\subsection{Validation experiments}
\label{sec:bias}

\subsubsection{Breast Cancer Data}
The goal of this experiment was to test if the CoLN combination method can compensate for distribution biases in the local models that create noise. For this, we used a feed forward Neural Network with 785 parameters (details of the architecture are explained in the supplementary material). 

In this experiment, we used breast cancer anonymized patient data from the University of Wisconsin Hospitals \cite{mangasarian1990cancer, wolberg1990multisurface, Dua:2019}.  We split the original data, which did not include gender information, into two ``virtual hospitals,'' \textit{Dataset1} and \textit{Dataset2}. Then we added to the original data an artificial gender parameter, in such a way as to introduce an opposite gender bias in cancer data for each virtual hospital. 
In \textit{Dataset1}, the data for patients with malignant cancer was predominantly labeled as ``male'', and data for patients with benign cancer was predominantly labeled as ``female''. Data in \textit{Dataset1} contains information from 230 patients with cancer, 115 malignant and 115 benign. Of the 115 malignant cancer patients 80\% were labeled as male and 20\%  as female.
In \textit{Dataset2}, the distribution is skewed in the other direction. It contains data from 354 other cancer patients,  177 malignant and 177 benign. Of the 177 malignant cancer patients, 80\% were labeled female and 20\% male. 
Test data was balanced, containing 136 samples, corresponding to 68 malignant cancer and 68 benign cancer, with gender being assigned randomly with  50\% for each gender in each class.

\begin{figure}[hbt!]
    \centering
    \includegraphics[width=6.0in]{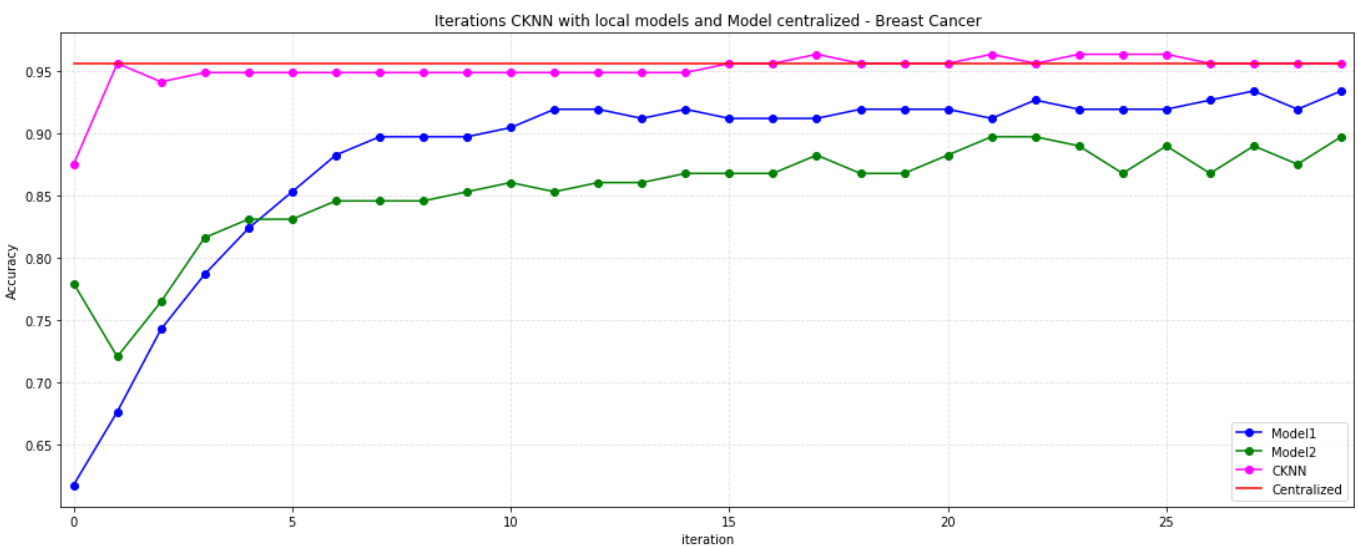}
    \caption{ Comparison of accuracy obtained in breast cancer data with distributional bias. Model1 identify models trained using Dataset1 and Model2 identify models trained using Dataset2.  In Round 0 the combined model used only the common initial random parameters,  Model1 and Model2 used parameters that resulted from the first round of training using local data. In subsequent rounds, Model1 and Model2 were the result of another round of training with local data  and the combined model used parameters combined in the previous round.}
    \label{fig:breastcancer}
 \end{figure}

We performed 30 rounds of combination.  \cref{fig:breastcancer} shows the results achieved by each model and by a centrally trained model using all 490 samples; test results for the centrally trained model were computed only once.  

The CoLN Combination reached  95.58\% accuracy in the second round, maintaining this value in the remaining rounds with few oscillations.  Local models \textit{Model1} and \textit{Model2} were initialized with the same random weights of the synchronizing model, with Model1 achieving initial accuracy 61.77\% and Model2 77.94\%.  Each local model at each round was trained using with 50 epochs. The centralized model achieved 95.59\% accuracy after 200 epochs. 
 
It is important to note that, even though,  after the first round, local models are initialized with the weights obtained from CoLN in the previous round (thus with a model of higher accuracy), after retraining with local data only, the result is a model of lower accuracy. Local models never achieve the same accuracy as CoLN.  This is expected as, at each round, the models are re-trained with biased data and incorporate the apparent dependency between gender and  cancer. However, CoLN combined learning model does not seem to be affected by this bias, and oscillates  with very little variation near its upper bound, which corresponds to almost the same accuracy as the centrally trained model. Detailed results for  each round are described in the supplementary material.

\subsubsection{Covid-19 Data}

In the second  experiment we used a covid-19 dataset consisting of 92 different clinical and laboratory tests from 5466 patients admitted to the Albert Einstein Hospital of São Paulo in 2020~\cite{mello_luiz_e_2020_3966427}. Clinical data from tests were used to predict whether the patient would be positive for covid-19, as mesured by a SARS-CoV-2 RT-PCR test. In the prediction task, a feed forward Neural Network was used containing 1,873 parameters. Details of the models are described in the supplementary material.

We randomly selected 811  patients for training and 588 for testing. The training set was randomly partitioned in two sub-sets with different distributions of positive and negative samples: Dataset1 consisted of  223 positive and 107 negative samples (total 330), and Dataset2 consisted  56 positive and 425 negative samples (total 481). 
The test dataset was balanced, with 279 positive and 279 negative cases. 

\begin{figure}[hbt!]
    \centering
    \includegraphics[width=6.0in]{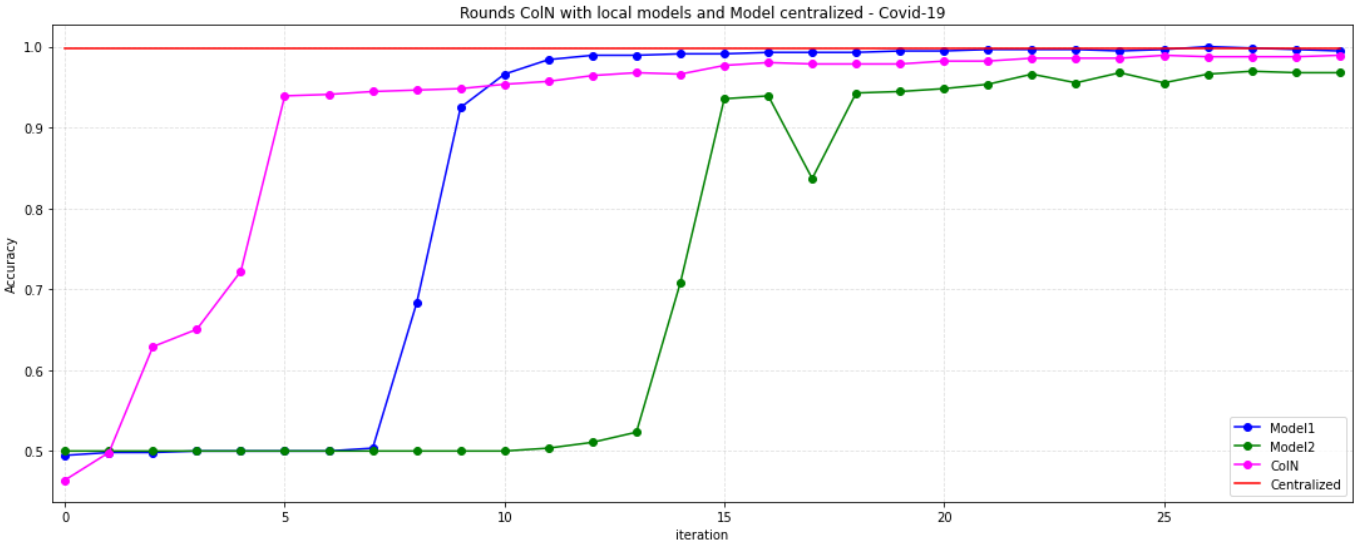}
    \caption{Comparison of accuracy obtained in COVID-19 data with distributional bias, in which CoLN reaches the 90\% accuracy in five rounds, and was followed by Model1 and Model2 in a few rounds. Model1 identify models trained using Dataset1 and Model2 identify models trained using Dataset2. In Round 0 the combined model used only the common initial random parameters,  Model1 and Model2 used parameters that resulted from the first round of training using local data. In subsequent rounds, Model1 and Model2 were the result of another round of training with local data and the combined model used parameters combined in the previous round. } 
        \label{fig:covid}
 \end{figure}

We used CoLN in 30 rounds of combination. Again, the  centralized model was trained only once. Results are shown in  \cref{fig:covid}. 

In this experiment, the centralized model achieved 100\% accuracy with 60 epochs. To demonstrate that CoLN can combine model features, the local training rounds were executed with only 10 epochs, and Model1 and Model2 achieved initial accuracy of  around 50\%, a result similar to randomly flipping a coin, and this behaviour was maintained in the next 7 rounds.  However, just after 5 rounds, CoLN already achieved more than 90\% accuracy, and the weights of this combined model were used  initial state for both Model1 and Model2.  Model1, whose local data was less unbalanced than Model2, took 10 combination rounds to achieve accuracy above 90\%, while Model2 took 15 rounds to reach the same level. This suggests that CoLN is not affected by unbalanced data, while local models struggle to overcome that deficit even when initialized with more accurate weights.  The results the each round are described in the supplementary material.

 \subsubsection{Text Sentiment Analysis}
 \label{sec:lstm}
 
 In the third experiment we used  used a Long-short term (LSTM) recurrent neural network  \cite{sak2014long}, for the problem of textual sentiment analysis, this time using five local hosts. Training for each host used as many epochs as necessary to achieve minimum loss.  
 The input consisted of a dataset of product reviews containing a numeric evaluation of products (1--5), an evaluation text and a binary-valued field indicating if the author of the review would recommend the product  to a friend~\cite{B2WReviews2019}. The neural network used the product review text to predict the recommendation.  We used a single-layer LSTM with a total of of 105,060 parameters. Details of the architecture are described in the supplementary material. 
 
\begin{figure}[hbt!]
    \centering
    \includegraphics[width=6.0in]{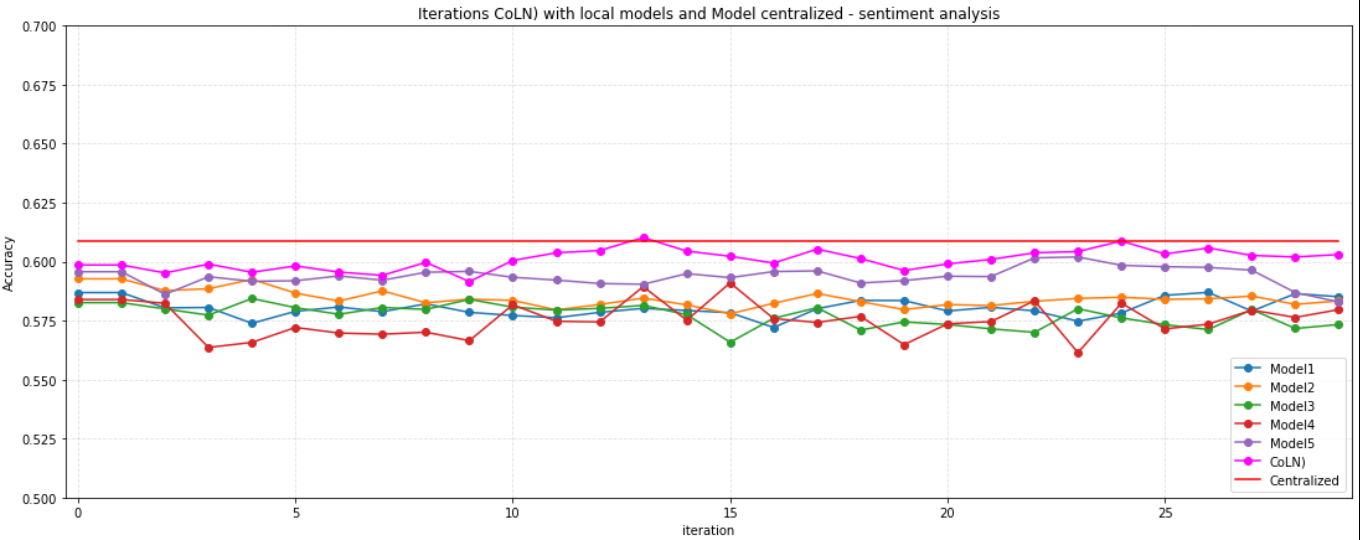}
    \caption{Comparison of  accuracy obtained for product review sentiment analysis employing LSTM models, with 5 distinct local hosts with distinct dataset sizes. CoLN reach stable accuracy in three rounds. Model1 identify models trained using Dataset1, Model2 identify models trained using Dataset2 and so forth until Model5.  In Round 0 the combined model used only the common initial random parameters,  Model1 to  Model5 used parameters that resulted from the first round of training using local data. In subsequent rounds, Model1 to Model5 were the result of another round of training with local data, but initializing the parameters with those obteind by the CoLN combination of the previous round, while the combination model results used the parameters from the combination in the same round. 
    }
    \label{fig:figlstm}
 \end{figure}

 The training set consists of 11890 evaluation texts and the testing set consisted of another 5096 evaluation texts. The training dataset was unevenly split among five local subsets one for each local model: Dataset1 (3304 texts), Dataset2 (2726 texts), Dataset3 (4436 texts), Dataset4 (4582 texts) and Dataset5 (1938 texts). The results are depicted in \cref{fig:figlstm}.

 Here the centralized model achieved 60.85\% accuracy after 180 epochs and the  CoLN combination reached 61.02\% accuracy in the 13th round.  Local models also stabilized, but lower accuracy levels (between 57 and 59\%).
 
 CoLN performed better than all local models, but not by much, as all models performed well, with differences mostly mirroring the sizes if each local dataset.

\subsubsection{Combining Convolutional Models for Image Classification}
\label{sec:cnn}

The goal of this experiment was to test the performance of the CoLN process on a convolutional neural network (CNN)~\cite{vedaldi2015matconvnet} in the task of multi-category image classification. We designed the experiment so that each of the two local hosts had unbalanced distributions of examples of the target  classification categories.  The classification task was performed by a CNN  with 1,658,700 parameters and an input matrix of 28x28x32. Details of the model are described in the supplementary material.

\begin{figure}[htb!]
    \centering
    \includegraphics[width=5.0in]{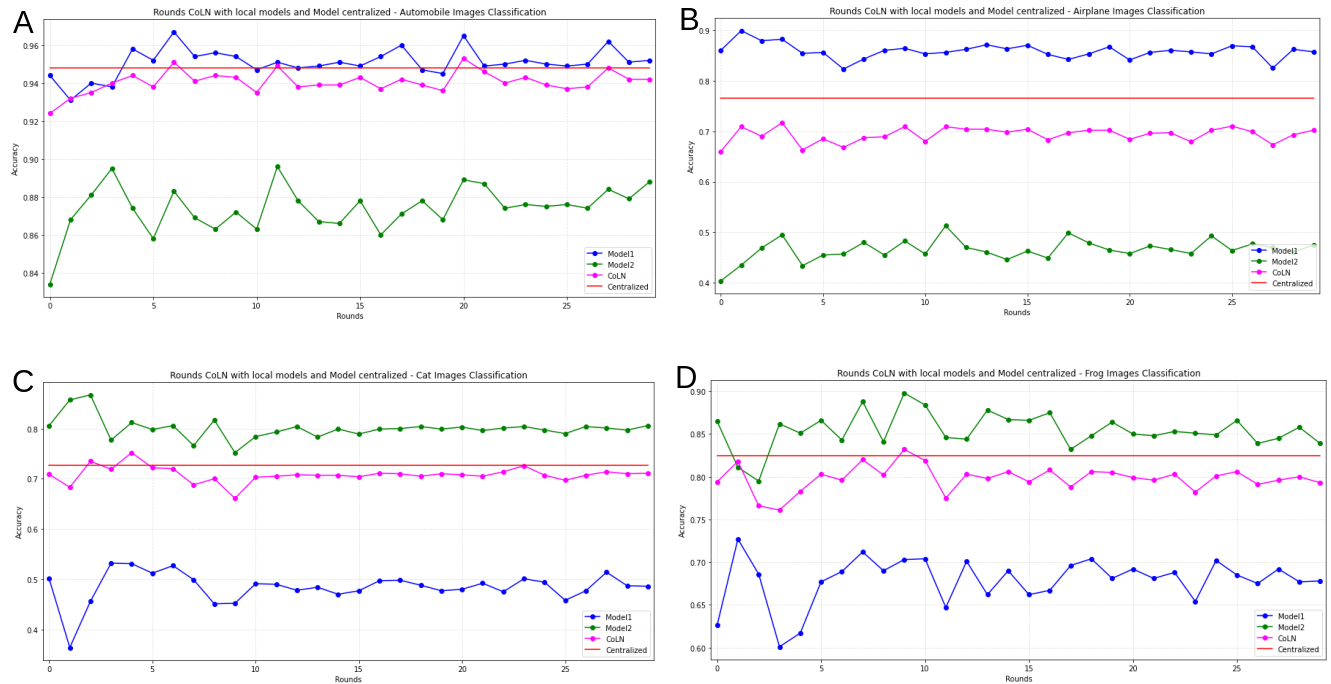}
    \caption{Comparison of classification accuracy for class A: ``Car'' , B: ``Airplane'' , C: ``Cat'' , D: ``Frog'' for image classification over CNN models over four image classes.    }
    \label{fig:cnnimageaut}
 \end{figure}

\begin{figure}[htb!]
    \centering
    \includegraphics[width=5.0in]{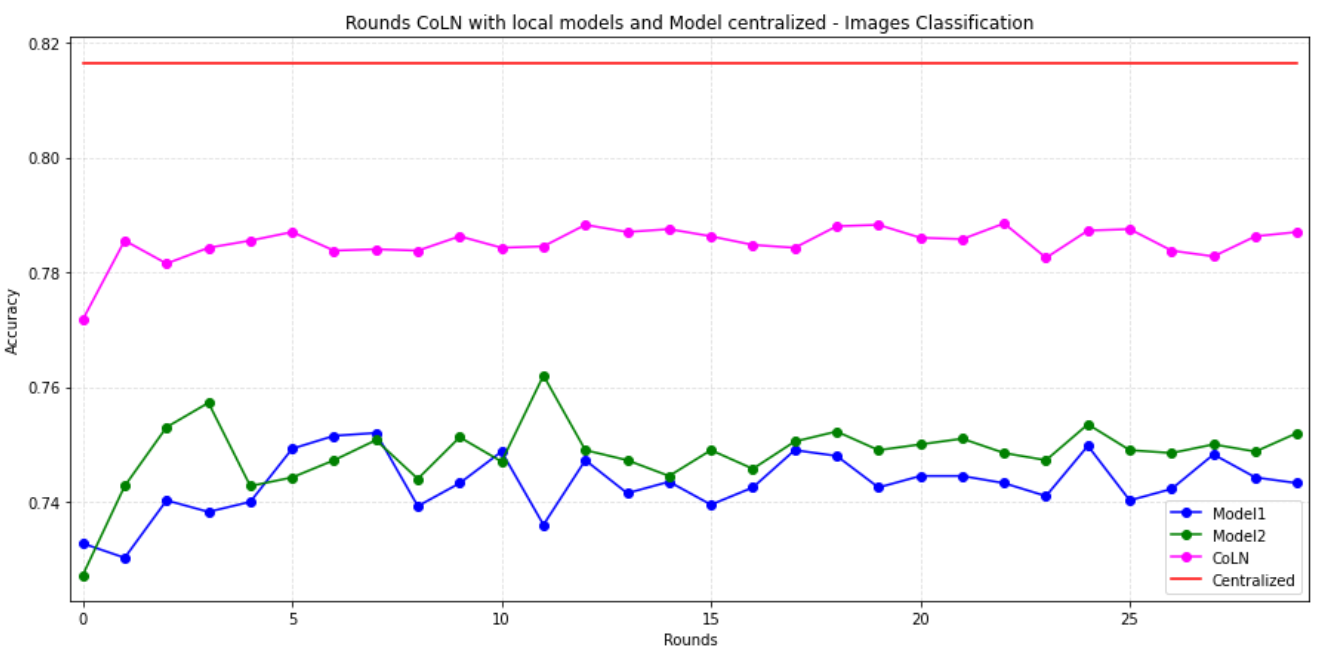}
    \caption{ Comparison of classification accuracy over all four classes of the  for image classification employing CNN models. Model1 was trained using Dataset1,  a set of images biased towards cats and dogs, Model2 was trained using Dataset2, a set of images biased towards planes and automobiles. In Round 0 the combined model used only the common initial random parameters,  Model1 and Model2 used parameters that resulted from the first round of training using local data. In subsequent rounds, Model1 and Model2 were the result of another round of training with local data and the combined model used parameters resulting from the CoLN combination of the previous round.}
    \label{fig:cnnimage}
 \end{figure}

We randomly selected 28,000 images from the CIFAR10 dataset \cite{krizhevsky2009learning} in four categories: cat, frog, car and airplane. Training images were separated in two local datasets: Dataset1 with the predominance of airplane and car images (4,000 airplane, 4,000 car, 2,000 frog, and 2,000 cat)  and Dataset2 with the predominance of frog and cat images (4,000 frog, 4,000 cat, 2,000 car and 2,000 airplane).  The test dataset  was a balanced set of 4,000 images, 1,000 of each category (airplane, car, cat and frog).
 We trained the centralized model only once with all 28,000 images of the combined training sets, and we performed   30 rounds of the  CoLN algorithm. Centralized training was peformed in 100 epochs and local training in 50 epochs.  The results  are shown in Figure \cref{fig:cnnimageaut}, and the combined accuracy in Figure \cref{fig:cnnimage}.

As we can see, Model1 is an ``expert'' in cars and airplanes and performs even better than the centralized model over these classes,  while Model2 performs poorly on theses categories, with a very low initial accuracy and only  a small improvement with CoLN.  Similarly, Model2 is an ``expert'' in Cats and Frogs, while Model1 performs poorly on those categories.
The CoLN combination,  with respect to each category, achieves an intermediary performance in relation to each ``expert'' model, but performs consistently below the centralized model over classes ``Airplane'' and ``Frog'' and achieves an accuracy close to that of the centralized model in classes ``Cat'' and ``Car''. 

The joint accuracy over all four classes is shown in \cref{fig:cnnimage}, revealing a picture more akin to previous results. The centralized  obtained the accuracy level of 81.65\%, and CoLN achieves 77.18\%, in the first round of combination, while local model's performance was 73.28\% and 72.73\%. CoLN obtains 78.85\% in the second round and oscillates around this value ($\pm0.03\%$) in the following rounds. The results show that clearly shows that CoLN ouperforms local models, approaching the ideal centralized performance.

\subsection{Comparing CoLN with Benchmark Federated Learning Methods over Image Classification}

\begin{figure}[htb!]
    \centering
    \includegraphics[width=5.0in]{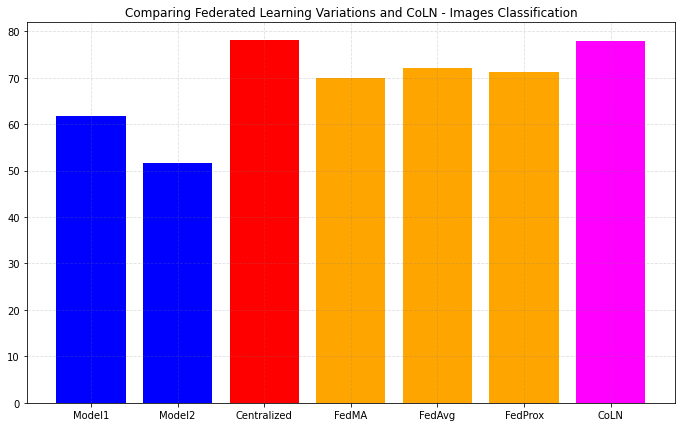}
    \caption{Comparison of combination methods over CIFAR10 image recognition task, employing two local hosts with fixed data. Five methods took part in the comparison: Centralised, FedMA, FedAvg, FedProx and CoLN. The bars indicate the final accuracy of each approach, the initial accuracy values for each of the local models, and the accuracy for the centralized model, trained with the combined local datasets. }

    \label{fig:cnnimagecom}
 \end{figure}

In this experiment we compared CoLN with with the federated learning methods using a Convolutional Neural Network Model in an image classification problem  using. For this, we used  GitHub implementation of FedMA, FedAvg, and FedProx available at \url{https://github.com/IBM/FedMA}, which was presented at FedMA presentation ~\cite{wang2020federated}.  
We downloaded from CIFAR10 60,000 32x32 color images from  of 10 different classes (airplanes, cars, birds, cats, deer, dogs, frogs, horses, ships, and trucks). From these, we created two  "local"  training sets with 25,000 images each and a test with 10,000 images (1,000 images of each category).  The centralized model was trained  with the combined training sets (50,000 images). The results are shown in  \cref{fig:cnnimagecom}. 

CoLN presented the best performance of all approaches. It achieved higher accuracy (77.8\%) when compared with FedMA (69.4\%), FedAvg (72.0\%) and FedProx (71.2\%) and its final accuracy was very close to that of the centralized model (77.8\% vs 78.0\%).
Also, CoLN'ns maximum accuracy of 77.8\% was reached in the second round, and other approaches in 6 rounds. All experiments were executed with 30 epochs and 6 rounds, and centralized in 120 epochs.

\newpage
\section{Conclusion}

Combined Learning of Neural network weights (CoLN) is a method that addresses privacy issues in neural network machine learning:  only model parameters and local dataset sizes are communicated, no data is ever exchanged.  It uses a efficient mathematical formulation to combine  a set of identical neural models whose parameters are autonomously trained over local data that is never shared. Only local model parameters and information on local dataset sizes are transmitted over the network by the collaborating hosts, maintaining privacy over individual data.  Our experiments in different Neural Network architectures have shown  that CoLN's combined models improve on local host performance, converges on small number of rounds, and   approximates the ideal situation where all data is available for training single, centralized model (from two to 6 steps, depending on the number of categories in the classification). CoLN also avoids several issues common to other distributed learning algorithms~\cite{li2018federated,mohri2019agnostic,wang2020federated}; it requires very little data traffic over the network, makes no assumption on data distribution,  and is resilient to local data size imbalance.

 Finally, CoLN compares favourably to other forms of federated learning models, exceeding their performance on a standard benchmark.

CoLN demonstrates a new direction for future work by combining the models
in a decentralized way using new mathematics capable of offering practical benefits
privacy, with security, as well as the ability to increase model accuracy
global without the need for each iteration to add new datasets and clients.
Experiments have shown that balanced learning can be a practical method
that allows collaborative research with datasets at each institution, mainly
especially when there is little data, significantly impacting research centers
smaller companies, start-ups and collaborative research across international borders.
Experiments and supplemental materials are available on the github website: \url{https://github.com/AlineIoste/CoLN} 

\bibliography{main}

\end{document}